%% file: main.tex
\documentclass{article}

 \usepackage[preprint,nonatbib]{neurips_2026}

\usepackage[utf8]{inputenc} 
\usepackage[T1]{fontenc}    
\usepackage{hyperref}       
\usepackage{url}            
\usepackage{booktabs}       
\usepackage{amsfonts}       
\usepackage{nicefrac}       
\usepackage{microtype}      
\usepackage{xcolor}         
\usepackage{csquotes}
\usepackage{mdframed}
\usepackage{tcolorbox} 

\usepackage{biblatex}
\usepackage[capitalize,nameinlink]{cleveref}

\definecolor{lightgray}{RGB}{240,240,240} 

\title{Out of Context: Reliability in Multimodal Anomaly Detection Requires Contextual Inference}

\author{
Kevin Wilkinghoff$^{1,2}$,
Neelu Madan$^{1,2}$,
Juan Miguel Valverde$^{3}$, Kamal Nasrollahi$^{1,4}$\\ \textbf{Radu Tudor Ionescu}$^{5}$, \textbf{Rafal Wisniewski}$^{1}$, \textbf{Thomas B. Moeslund}$^{1,2}$, \textbf{Wenwu Wang}$^{6}$\\ \textbf{Zheng-Hua Tan}$^{1,2}$\\
$^{1}$Aalborg University, Denmark \quad
$^{2}$Pioneer Centre for Artificial Intelligence, Denmark \\
$^{3}$Technical University of Denmark, Denmark \quad
$^{4}$Milestone Systems, Denmark \\
$^{5}$University of Bucharest, Romania \quad
$^{6}$University of Surrey, UK
}

\begin{document}

\maketitle

\begin{abstract}
Anomaly detection aims to identify observations that deviate from expected behavior. Because anomalous events are inherently sparse, most frameworks are trained exclusively on normal data to learn a single reference model of normality. This implicitly assumes that normal behavior can be captured by a single, unconditional reference distribution. In practice, however, anomalies are often context-dependent: A specific observation may be normal under one operating condition, yet anomalous under another. As machine learning systems are deployed in dynamic and heterogeneous environments, these fixed-context assumptions introduce structural ambiguity, i.e., the inability to distinguish contextual variation from genuine abnormality under marginal modeling, leading to unstable performance and unreliable anomaly assessments. While modern sensing systems frequently collect multimodal data capturing complementary aspects of both system behavior and operating conditions, existing methods treat all data streams equally, without distinguishing contextual information from anomaly-relevant signals. As a result, abnormality is often evaluated without explicitly conditioning on operating conditions. We argue that \textbf{multimodal anomaly detection should be reframed as a cross-modal contextual inference problem, in which modalities play asymmetric roles, separating context from observation, to define abnormality conditionally rather than relative to a single global reference}. This perspective has implications for model design, evaluation protocols, and benchmark construction, and outline open research challenges toward robust, context-aware multimodal anomaly detection.
\end{abstract}

\section{Introduction}

Anomaly detection aims to identify observations that deviate from expected or normal behavior. What constitutes abnormality depends on the perspective adopted. In probabilistic terms, anomalies are often associated with low-probability events or distribution tails, while in other settings they may be defined relative to learned patterns, structural deviations, or human perception. Regardless of the formulation, what is considered abnormal is often context-dependent: a pattern that signals failure in one situation may be entirely expected in another. For example, an elevated heart rate is normal during physical exercise, but may indicate a medical emergency during rest \cite{shaffer2017overview}. In anomaly detection, such cases are referred to as \emph{contextual anomalies} \cite{chandola2009anomaly,pang2022deep,ruff2021unifying}. Similar ambiguities arise in industrial monitoring, autonomous systems, and video surveillance, where identical observations may have different interpretations under different operating conditions. In visual domains, context frequently arises from spatial and semantic structures that characterize normality \cite{ristea2022self,madan2023self}, and in monitoring applications it may be temporal, as gradual wear or evolving acoustic signatures are expected over time, while deviations from these patterns indicate anomalous behavior \cite{audibert2020usad,zhao2019deep}. In realistic deployments, systems operate under evolving and heterogeneous contextual conditions, and  behavior can shift across operating regimes without necessarily indicating genuine anomalies \cite{moreno-torres2012unifying,sugiyama2012machine}. Nevertheless, most anomaly detection methods implicitly assume that the conditions that define normal behavior remain unchanged. As a result, \textbf{anomaly detection systems must distinguish genuine abnormal behavior from contextual variation, a distinction that static and context-independent reference distributions are poorly suited to capture} \cite{ruff2021unifying,snoek2019can}.

 Many anomaly detection systems \cite{georgescu2021anomaly, ristea2022self, madan2023self, madan2021temporal, wilkinghoff2025local} are trained using only normal data, as anomalous events are rare and difficult to collect in sufficient quantity. This scarcity affects both training and evaluation: constructing realistic anomalous test data is highly application-dependent and limited, allocating them for training further reduces the amount of reliable evaluation data. Consequently, anomaly detection systems trained exclusively on normal data, commonly referred to as semi-supervised anomaly detection \cite{aggarwal2017outlier}, implicitly assume that the contextual conditions observed during training remain representative of those encountered during deployment. In practice, however, normal behavior varies across environmental, operational, and behavioral contexts. Effective anomaly detection, therefore, requires not only modeling normal behavior but also distinguishing deviations caused by contextual variation from those caused by genuinely abnormal events. These challenges are further amplified when systems rely on domain-specific training or tuning, as adapting to new operating conditions can become costly and impractical if large amounts of context-specific data are required. Therefore, \textbf{anomaly detection systems should be designed to remain flexible and capable of adapting to evolving contextual conditions without extensive retraining or manual reconfiguration, ideally through models that explicitly capture contextual structure}.

Capturing contextual information is central to achieving reliable and adaptable anomaly detection. Multimodal sensing provides a natural pathway toward this goal and is increasingly common in domains such as healthcare, industrial monitoring, and autonomous systems, where multiple data sources capture complementary aspects of system behavior \cite{feng2020deep,krones2025review}. However, multimodal machine learning is traditionally formulated through joint representation learning and feature fusion, treating modalities as symmetric and interchangeable inputs \cite{baltrusaitis2019multimodal,ngiam2011multimodal,radford2021learning,ramachandram2017deep}. 
When these paradigms are adopted in anomaly detection, multimodal observations are typically incorporated through feature fusion or cross-modal consistency objectives \cite{constanzino2023multimodal,park2018multimodal,park2016multimodal,sipple2020interpretable,wang2023multimodal}, producing joint anomaly scores over the combined feature space rather than explicitly conditioning on contextual variables.
This symmetric treatment entangles feature representations, limiting the ability of multimodal systems to express contextual roles and obscuring the interpretation of anomaly decisions \cite{traeuble2021disentangled}. As a result, multimodal systems inherit key limitations of unimodal anomaly detection, particularly the difficulty of separating contextual variability from genuine anomalies, thereby restricting reliability and generalization in complex environments. Overcoming these limitations requires moving beyond feature-level fusion toward explicit conditional modeling of context. We argue that \textbf{anomaly detection should be reframed as a cross-modal contextual inference problem, in which modalities play asymmetric roles and abnormality is defined conditionally rather than relative to a single global reference}. Here, contextual inference refers to inferring the operating conditions under which observations are generated and assessing abnormality relative to those conditions. This reframing has direct implications for model design, evaluation, and interpretation.


\section{Why Fixed-Context Anomaly Detection Fails}
\label{sec:failure}

Most anomaly detection methods assume that normal data follow a single, stable distribution. In practice, a model is trained on data collected under certain operating conditions and learns a reference distribution $p(x)$ that represents “normal” behavior. Decision thresholds are then calibrated using this training data and assumed to remain valid at deployment.

In real-world systems, however, normal behavior rarely arises from a single fixed regime. Instead, observations are generated under multiple contextual conditions $c \in \mathcal{C}$, such as different environments, operating modes, or user states. Each context induces its own conditional distribution $p(x \mid c)$. If contextual information is ignored, these distinct distributions are symmetrically aggregated into a single marginal reference,
\begin{equation}
    p(x) = \int p(x \mid c)\, p(c)\, dc,
\end{equation}
which treats all contextual states equivalently rather than conditioning on them as in classical mixture-of-experts models \cite{jacobs1991adaptive,jordan1994hierarchical}. As illustrated in \cref{fig:context_vs_marginal}, marginalization mixes variation from contextual changes with variation from genuine anomalies, blurring the distinction between the two.

\textbf{Failure mode.} This symmetric treatment of contextual regimes (i.e., ignoring the conditioning variable) induces a predictable failure mode. An observation $x$ unlikely under the correct context-specific distribution $p(x \mid c_2)$ (as shown in Fig. \ref{fig:context_vs_marginal}) may fall in a high-density region of the marginal $p(x)$ and appear normal when context is ignored, while observations typical in one regime may be erroneously flagged in another. A model optimized for density estimation under the mixture distribution $p(x)$ is therefore not optimized for detecting conditional tail events. Without an explicit conditioning structure, such tail events are generally not identifiable from the marginal alone, since multiple conditional decompositions may induce the same $p(x)$. The problem is thus miscalibration of the marginal density with respect to context-specific tail events, rather than a failure of marginal modeling itself. This limitation persists even with highly expressive components, since maximizing marginal likelihood does not enforce alignment with the underlying, semantically meaningful contextual regimes. More importantly, the field typically models anomaly detection marginally, while abnormality in heterogeneous environments is inherently defined conditionally. Unless contextual structure is observed or inferred, the data alone cannot determine which variations reflect regime shifts and which reflect genuine abnormality.

\textbf{Implications for evaluation.} This limitation is not restricted to explicitly threshold-dependent metrics. Common evaluation measures in anomaly detection, including the area under the ROC curve (AUC) \cite{fawcett2006introduction} and partial AUC \cite{mcclish1989analyzing}, are formally threshold-independent but still rely on a globally consistent ranking of $p(x)$. When contextual shifts alter the relative ordering of normal and anomalous observations, performance can degrade even under such metrics, since dataset shift disrupts score separability and ranking consistency \cite{bendale2016towards,hendrycks2017baseline,salehi2022unified}. Thus, even threshold-independent evaluation assumes a marginal ranking, implicitly treating it as context-invariant. Ignoring context can therefore degrade both calibration and ranking, because rarity under the marginal distribution is not equivalent to conditional abnormality. An observation may be infrequent overall yet entirely expected within a specific operating regime.
Consequently, prevailing objectives and metrics remain aligned with marginal density estimation rather than with context-conditional detection, highlighting that current objectives and benchmarks favor marginal ranking over context-conditional abnormality.

\input{figures/fail_illustration}

\textbf{Relation to distribution shift.} This issue is related to distribution shift, but differs in an important way. Domain adaptation and generalization methods aim to reduce the distribution mismatch between source and target domains to preserve predictive performance \cite{kouw2021review,yang2024can,zhou2023domain}. In anomaly detection, however, distributional differences often define what is considered normal rather than representing nuisance variation. Because contextual variables directly influence anomaly semantics, enforcing unconditional invariance may remove precisely the structure needed to distinguish contextual variability from abnormal behavior. The goal is not to remove distributional variation through invariance, but to model how normal behavior varies across contexts and how this variation defines abnormality.
Recent work has begun to address distribution shift in anomaly detection \cite{cao2023anomaly,carvalho2023invariant,wilkinghoff2025handling,wilkinghoff2025local}, but existing approaches remain limited. Many operate in single-modality settings, assume predefined contextual partitions, or rely on static domain definitions. Such assumptions are difficult to maintain in environments where contextual conditions are continuous, partially observed, or dynamically evolving. A more appropriate formulation is to model context explicitly and condition anomaly assessment on it.

In summary, the fundamental limitation of fixed-context anomaly detection is not the deviation scoring mechanism itself, but the use of an unconditional and symmetrized reference distribution that collapses distinct contextual regimes into a single marginal model. As a result, contextual variability and abnormal behavior become indistinguishable, complicating reliable evaluation and deployment.


\section{Position Statement}
\label{sec:position}

The structural ambiguity illustrated above motivates reformulation rather than incremental adjustment. If abnormality is defined relative to operating conditions, anomaly detection must be formulated as conditional inference over structured contextual variables, evaluating whether an observation is unlikely under the conditional distribution $p(x \mid c)$. Under this view, abnormality is not an intrinsic property of observations but a relational property defined relative to operating conditions. Without structural constraints tying latent variables to contextual semantics, conditioning remains statistically, but not semantically, identifiable.

\input{figures/context_example}
\textbf{Taxonomy of context.}
We distinguish three complementary forms of context:
(i) \emph{environmental context}, referring to externally defined conditions under which observations are generated (e.g., operating regimes or temporal phases), including both directly observed variables and deterministic transformations thereof (e.g., operational state derived from time and policy);
(ii) \emph{latent context}, capturing unobserved factors that induce structured variability in the data distribution;
and (iii) \emph{modal context}, where contextual information is encoded in auxiliary modalities that are directly observed and used to infer context.
These forms differ in observability and representation, but all condition the distribution of primary signals.
An illustrative example is shown in \cref{fig:context_example}, where these different forms of context jointly determine the interpretation of identical observations.

\textbf{Multimodal perspective.} Modal context provides a direct, observable proxy, whereas environmental and latent context are not directly accessible and must be inferred or specified externally. Certain modalities capture primary signals of interest, while others encode contextual evidence that governs their interpretation. Crucially, modalities need not play symmetric roles. Treating multimodal inputs as interchangeable features reproduces the symmetric aggregation that causes marginal failure. Instead, multimodal anomaly detection should be understood as inherently asymmetric, where auxiliary modalities provide contextual information that shapes the interpretation of primary signals, consistent with the conditional structure $p(x \mid c)$.

Physiological monitoring illustrates this asymmetry. Heart rate measurements serve as the primary signal, while accelerometer or temperature data encode contextual evidence about activity or environment. Elevated heart rate may be anomalous at rest but expected during exercise. Contextual modalities support interpretation of cardiac dynamics, yet abnormal cardiac events cannot be inferred reliably from contextual signals alone. This example illustrates anomaly detection as conditional inference over primary signals given contextual evidence.

A similar asymmetry arises in video surveillance. An individual running may indicate anomalous behavior in a restricted area or during nighttime, but is entirely expected in a public space during daytime. Visual motion patterns constitute the primary signal, while contextual information such as scene type, time, or activity patterns governs their interpretation. Without conditioning on such context, identical observations may be assigned inconsistent anomaly scores.

\textbf{Interpretability.} Beyond statistical robustness, explicit contextual modeling also improves interpretability \cite{doshi2017towards,rudin2019stop, singh2023eval}. When anomaly scores are conditioned on identifiable contextual variables rather than produced by undifferentiated representations, it becomes clearer why an observation is considered abnormal. Context provides an explanation: an observation is anomalous relative to specific operating conditions. This transparency is crucial in safety-critical and monitoring applications, where anomaly decisions must be understood and trusted by human operators \cite{rudin2019stop,schoelkopf2021toward}.

\begin{tcolorbox}[
  colback=lightgray,
  colframe=black,
  boxrule=0.8pt,
  arc=3pt,
  left=6pt,
  right=6pt,
  top=1.25pt,
  bottom=1.25pt
]
\textbf{Position:} Anomaly detection should be formulated as conditional inference over contextual variables, allowing multimodal observations to play explicit and potentially asymmetric roles.
\end{tcolorbox}

This perspective extends beyond model design. It requires rethinking evaluation practices, benchmark construction, and the interpretation of results. We therefore advocate treating contextual inference as a first-class modeling objective rather than as an implicit by-product of representation learning.


\section{Contextual Anomaly Detection as Conditional Inference}
\label{sec:formalization}

We now formalize anomaly detection as conditional inference over contextual variables.
Let $x \in \mathcal{X}$ denote an observation and $c \in \mathcal{C}$ its contextual state under which it is generated, such as operating conditions, auxiliary signals, system modes, or other sources of prior information. 
An observation is anomalous if it is unlikely under the conditional distribution $p(x \mid c)$.
When $c$ is latent or partially observed, anomaly detection requires inference under uncertain context.

\textbf{Inference vs.\ aggregation.} Importantly, simply evaluating $x$ under multiple conditional distributions does not resolve this ambiguity. 
Taking a minimum or maximum over $\{p(x \mid c)\}_{c \in \mathcal{C}}$ either masks anomalies by selecting the most compatible context or exaggerates deviations by selecting the least compatible one. 
Such aggregation strategies retain the symmetric treatment of contextual regimes and therefore reproduce the same entanglement illustrated in \cref{fig:context_vs_marginal}. 
Moreover, in realistic settings the correct contextual state is rarely known a priori, making manual selection of $c$ impractical or ill-defined. 
Conditioning must therefore be achieved through inference rather than enumeration.
This reframing connects contextual anomaly detection to broader themes in probabilistic machine learning, including latent variable modeling \cite{kingma2014auto}, representation learning \cite{bengio2013representation}, conditional density estimation \cite{papamakarios2021normalizing}, and conditional generative modeling \cite{sohn2015learning}. In these settings, models aim to separate variability induced by structured factors from meaningful deviations \cite{locatello2019challenging}.

\textbf{Multimodal inference.} Building on this perspective, multimodal observations provide a principled mechanism for contextual inference in practice. 
Certain modalities represent primary signals whose behavior is evaluated for abnormality, while others provide contextual evidence that governs how those primary signals should be interpreted.
This asymmetry can be learned implicitly from data, for example through architectures in which auxiliary modalities condition or modulate representations of primary signals.
This enables implicit context inference without manual contextual partitioning.

Let $\mathcal{M}$ denote a finite index set of modalities and $x := (x_m)_{m \in \mathcal{M}}$ a multimodal observation with $x_m \in \mathcal{X}_m$.
Anomaly detection can then be expressed as conditional evaluation of a subset of modalities given the others,
\begin{equation}
    x_S \not\sim p\bigl(x_S \mid x_{\mathcal{M}\setminus S}\bigr),  \qquad S \subseteq \mathcal{M}.
\end{equation}
This formulation makes explicit that modalities may serve as contextual variables for one another and their roles are inherently asymmetric. Anomalies are detected by assessing whether a primary modality behaves consistently with expectations induced by evidence from auxiliary modalities.

This probabilistic view clarifies that contextual anomaly detection is a structured inference problem. 
Effective systems must (i) learn representations that capture contextual structure, (ii) infer contextual state at deployment time, and (iii) evaluate deviations relative to this inferred context, e.g., by inferring context from auxiliary modalities and conditioning the evaluation of primary signals on this estimate.
Progress may therefore depend as much on advances in \textbf{context-aware representation learning} and \textbf{structured inference} as on improved scoring mechanisms.

\textbf{Relation to open-set recognition.} The conditional formulation also connects anomaly detection to open-set classification \cite{geng2021recent,scheirer2013toward}, where class labels $y \in \mathcal{Y}$ define discrete contextual states (i.e., $c = y$) and abnormality is assessed via $p(x \mid y)$. Open-set recognition can thus be viewed as a supervised special case of contextual anomaly detection. In many deployment settings, however, contextual variables are continuous, multimodal, and partially observed, extending conditional abnormality beyond fixed and fully supervised regimes.

Finally, we do not claim that multimodal data are universally required. Contextual structure can often be inferred from a single modality using temporal, spectral, or structural features. Rather, anomaly detection becomes ill-posed without a conditioning structure whenever normal behavior varies across regimes. Multimodal observations, when available, provide a natural mechanism for representing such structure by separating primary signals from contextual evidence. The central contribution is therefore not the inclusion of additional modalities per se, but the recognition that anomaly detection is incomplete without explicit reasoning about the conditions under which observations are generated.


\section{Competing Perspectives and Rebuttals}

Our position that anomaly detection requires conditional inference over structured contextual variables challenges prevailing assumptions in anomaly detection and multimodal learning, particularly the adequacy of symmetric and unconditional modeling strategies. It questions whether symmetric representation learning \cite{baltrusaitis2019multimodal}, large-scale implicit modeling \cite{radford2021learning}, or unconditional reference distributions augmented through heuristic adaptation \cite{hendrycks2017baseline} are sufficient to resolve contextual ambiguity. 

\paragraph{Counter-Argument 1: Joint representation learning implicitly captures context.}

A common view is that multimodal representation learning captures contextual relationships implicitly through shared latent embeddings across modalities \cite{baltrusaitis2019multimodal,ngiam2011multimodal}. Because cross-modal dependencies are encoded with a unified representation, explicit contextual modeling is considered unnecessary.

However, the central failure of fixed-context anomaly detection arises from symmetric aggregation of heterogeneous regimes into an unconditional reference distribution $p(x)$. Joint embeddings do not prevent this unless modalities are assigned explicit roles. Even approaches such as temporal conditioning \cite{madan2021temporal}, which capture contextual dependencies, do not guarantee preservation of conditional semantics in the learned representation. In effect, symmetric embedding spaces implement marginalization implicitly by collapsing contextual regimes into a shared representation. Shared representations can therefore entangle contextual variability with behavior that should be evaluated for abnormality, reproducing the statistical ambiguity illustrated in \cref{fig:context_vs_marginal}. Moreover, joint embeddings are typically optimized for predictive coherence or reconstruction fidelity rather than for preserving conditional semantics \cite{locatello2019challenging}. They do not guarantee separation between variables that define operating conditions and variables evaluated relative to those conditions. In contrast, the conditional formulation explicitly mirrors the structure $p(x \mid c)$ and preserves the asymmetric roles central to our position.

\paragraph{Counter-Argument 2: Sufficiently large or generative models can learn context implicitly.}

Another perspective holds that sufficiently large-scale expressive or generative models may learn contextual dependencies directly from data, even without explicit contextual supervision \cite{alayrac2022flamingo,neyshabur2020role}. If contextual structure is present in the data, scale and expressivity should uncover it.

Although highly expressive generative models can capture complex cross-modal correlations, anomaly detection operates in an asymmetric regime where anomalous events are rare or absent during training. Context must therefore be inferred solely from normal data. However, learning a marginal distribution $p(x)$ is not the same as detecting deviations relative to a contextual distribution $p(x \mid c)$. A model can estimate the overall data distribution accurately and still fail to distinguish contextual variation from genuine anomalies. Increasing model capacity does not resolve this issue, because the learning objective remains unconditional. Moreover, when contextual cues are missing, corrupted, or shifting, models without explicit conditioning lack a principled way to reason about contextual uncertainty. Explicit conditional modeling instead separates contextual inference from deviation scoring and supports more reliable anomaly assessment.

\paragraph{Counter-Argument 3: Explicit contextual modeling introduces unnecessary complexity or requires regime-specific data.}

A common objection is that modeling contextual variables increases architectural and computational complexity, or that it requires collecting data for each operating regime and maintaining multiple reference distributions \cite{chandola2009anomaly,pang2022deep,wilkinghoff2026much}. From this perspective, unconditional models appear simpler and more practical because they avoid explicit regime partitioning.

However, ignoring context does not eliminate complexity but instead displaces it into hidden failure modes. Fixed-context anomaly detection systems frequently require manual threshold adjustments, domain-specific retraining, or heuristic recalibration when deployed under distributional or contextual shifts \cite{hendrycks2017baseline,moreno-torres2012unifying,pang2022deep,salehi2022unified,tack2020csi}. These interventions reflect an implicit and reactive form of contextual inference. They approximate contextual reasoning in an ad hoc manner without exposing the underlying dependencies.

Moreover, exposure to contextual variability is necessary for reliable anomaly detection regardless of modeling strategy. Without it, any model, conditional or unconditional, will interpret contextual shifts as anomalous behavior. Marginal strategies must either symmetrically pool heterogeneous regimes, thereby reintroducing statistical entanglement as illustrated in \cref{fig:context_vs_marginal}, or maintain separate unconditional models for each regime without shared structure. In contrast, the conditional framework enables inference over contextual state rather than manual enumeration of regimes. Multimodal observations provide a practical mechanism for this inference by encoding contextual evidence that conditions the primary signals, thus preserving the asymmetric structure articulated in our position.


\section{Multimodal Context as the Next Frontier in Anomaly Detection}

Multimodal anomaly detection remains an emerging area without consolidated benchmarks or consistent evaluation paradigms, underscoring the need for clearer principles for defining and assessing abnormality under heterogeneous regimes. Representing context, defining abnormality, structuring inference, and designing evaluation protocols are tightly coupled requirements for preserving conditional semantics. We outline central challenges below.

\paragraph{How should contextual variables be defined, represented, and identified?}

A key challenge is how to represent and identify contextual variables within the taxonomy introduced in \cref{sec:position} while preserving their conditional role relative to primary signals.

As shown in the literature on disentangled and causal representation learning \cite{locatello2019challenging,schoelkopf2021toward}, latent variables are generally non-identifiable without structural assumptions or inductive bias. Contextual variables should therefore not simply appear as arbitrary latent dimensions that can be reparameterized without changing the marginal distribution. Instead, they must correspond to structured factors that meaningfully influence primary behavior. This raises fundamental design questions: should context be modeled as discrete operating regimes, continuous latent states, hierarchical structures, causal drivers, or hybrid representations that combine symbolic and sensory information? Without principled representational constraints, conditional modeling may merely relocate contextual entanglement into latent space rather than resolving it.

\paragraph{Are anomalies inherently context-dependent or context-invariant?}

Not all anomalies depend equally on context. Some deviations may violate global physical constraints or safety conditions and thus manifest across operating regimes. Others emerge only relative to specific contextual conditions.

This suggests a taxonomy of anomalies that distinguishes between context-invariant violations and context-relative deviations. Clarifying this distinction is essential for determining when conditional modeling is necessary and when marginal modeling is not only sufficient but preferable for its simplicity. It also has implications for generalization: models optimized for context-relative anomalies may not capture global constraint violations, and vice versa. Existing taxonomies distinguish between point and contextual anomalies \cite{chandola2009anomaly}, but do not provide criteria for when unconditional deviation detection suffices and when conditional modeling is required.

\paragraph{How should conditional inference and anomaly scoring be structured?}

A central challenge is that contextual inference and anomaly scoring cannot be collapsed into a single undifferentiated representation. Models must therefore be structured so that contextual state is inferred explicitly and anomaly scores are computed relative to that state, rather than allowing contextual variability to be absorbed into primary representations. 

This raises questions about how conditional inference and deviation evaluation should be factored within a unified probabilistic model. Should contextual state be inferred first and anomaly scoring performed conditionally, or should both be optimized jointly? How should model objectives enforce asymmetric roles between contextual and primary variables? Architectural choices such as latent variable models \cite{kingma2014auto}, conditional generative models \cite{rahman2024conditional}, graphical models \cite{shrivastava2023neural}, causal \cite{schoelkopf2021toward} formulations, or approaches based on disentangled latent representations \cite{higgins2017beta-vae} impose different constraints on how context governs observable behavior.

\paragraph{When and how does multimodality resolve contextual ambiguity?}

Multimodal sensing provides a natural mechanism for contextual inference when contextual variables are not directly observable. However, incorporating additional modalities introduces trade-offs. Additional modalities may improve contextual identifiability but may also introduce noise, misalignment, or new failure modes.

The critical question is whether multimodality preserves conditional structure and asymmetric roles. Modality asymmetry may need to be imposed architecturally, inferred dynamically, or enforced through objective constraints that encode conditional semantics. Understanding how modality roles should be defined, learned, or adapted over time is central to avoiding symmetric aggregation in multimodal settings. Without explicit role differentiation, multimodal systems risk collapsing into joint representations that reproduce marginal entanglement.

\paragraph{How should contextual anomaly detection be evaluated?}

Evaluation protocols that fail to separate these effects risk rewarding models for exploiting marginal correlations rather than performing genuine contextual reasoning. Most existing anomaly detection benchmarks are unimodal and do not explicitly model systematic context shifts across operating regimes \cite{acsintoae2022ubnormal,ahmad2017unsupervised,bergmann2022beyond,bergmann2019mvtec,koizumi2020description}. While several benchmarks introduce domain or contextual shifts \cite{dohi2023description,dohi2022description,kawaguchi2021description,sultani2018real,liu2018future,nishida2024description,nishida2025description}, their evaluation protocols typically rely on single source–target domain splits, limiting systematic assessment of conditional abnormality across multiple contextual regimes. Multimodal anomaly detection datasets remain comparatively rare \cite{albertini2024imadds,bergmann2022mvtec3dad,jiang2025mmad} and often inherit the same evaluation limitations.

Developing principled evaluation protocols therefore requires isolating conditional abnormality from distributional robustness. Evaluation should test whether models detect anomalies relative to inferred context, generalize across regimes, and remain reliable under missing or corrupted evidence.

\paragraph{How should anomaly detection operate under evolving contextual dynamics?}

Real-world systems operate under continuously evolving conditions in which contextual variables change over time. If contextual state governs the distribution of primary signals, then anomaly detection becomes a problem of inference over dynamical context. Context may evolve gradually, shift abruptly, or exhibit hierarchical structure across timescales.

This perspective connects contextual anomaly detection to continual learning \cite{delange2022continual, doshi2020continual}, domain adaptation \cite{zhou2023domain}, and online inference \cite{canini2009online}. However, existing approaches rarely treat contextual inference as a primary design objective. A conditional framework requires models that can update contextual beliefs, propagate uncertainty across time, and distinguish between contextual evolution and genuine system failures. This shifts anomaly detection from static deviation detection toward adaptive probabilistic reasoning under  evolving and nonstationary conditions.

\paragraph{Can contextual variables themselves be anomalous?}

Deviations in context may themselves constitute anomalies. This suggests a hierarchical anomaly detection problem in which anomalies may occur at multiple levels: signal-level deviations conditioned on context, context-level deviations that alter the governing distribution, or joint violations involving both signal and context dynamics.

Recognizing this hierarchy expands anomaly detection beyond single-level scoring and raises questions about multi-level inference, causal relationships between context and signal, and how anomalies propagate between them. It also improves interpretability by clarifying whether abnormality arises from signal-level deviations, contextual shifts, or their interaction. Addressing these challenges requires models that reason jointly over signal and context rather than treating context as static.

These considerations motivate the following central challenges.

\paragraph{Grand Challenge 1: Toward a Theory of Conditional Abnormality.}

The preceding questions point to a deeper theoretical gap: a principled theory of conditional abnormality. Such a theory must characterize contextual variables, characterize how they govern normal behavior, and formalize the conditions under which marginal and conditional anomaly scores diverge. It should formalize when anomaly detection is well-posed under conditional modeling and when context-independent modeling is insufficient. Establishing this foundation would unify anomaly detection with structured probabilistic inference and enable formal guarantees of robustness under contextual shifts.

\paragraph{Grand Challenge 2: Context-Aware Anomaly Detection Systems That Reason Under Uncertainty.}

A second challenge is designing anomaly detection systems that explicitly reason about contextual state, its uncertainty, and its evolution over time. This requires representing uncertainty over context, propagating contextual uncertainty into anomaly decisions, and establishing principled decision rules when both signal and contextual state are uncertain. Achieving this goal would transform anomaly detection from static deviation scoring into adaptive decision-making under uncertainty. Such systems would be capable of operating reliably in complex environments where contextual information is incomplete, conflicting, or continuously changing.

\section{Conclusion}

Anomaly detection is contextual if what counts as abnormal depends on the operating conditions. The same observation may be normal in one regime and anomalous in another, yet most existing approaches model a single unconditional reference distribution that aggregates heterogeneous regimes symmetrically. As argued, this marginal treatment introduces structural ambiguity: contextual variability and genuine abnormality become statistically entangled in a way that is not resolved by increased model capacity under unconditional objectives.

We argue that anomaly detection should be formulated as conditional inference over structured contextual variables. Multimodal data provide a natural mechanism for this perspective, as auxiliary modalities encode contextual evidence that governs the interpretation of primary signals. Crucially, modalities need not play the same role. By explicitly modeling how contextual variables condition primary observations, anomaly detection shifts from unconditional deviation detection toward principled probabilistic inference under inferred operating conditions.

More broadly, this reframing presents anomaly detection as a problem of context-aware representation learning and structured probabilistic inference. It motivates models that preserve conditional semantics, evaluation protocols that separate contextual variation from abnormality, and theoretical frameworks that clarify when anomaly detection is well-posed. Treating context as a first-class modeling component is therefore an architectural preference, not a correction for a deeper problem-definition mismatch in which abnormality is modeled and evaluated marginally, despite being defined conditionally in heterogeneous environments.

\section{Generative AI disclosure}
Generative AI tools were used for language editing and polishing of the manuscript.
All scientific content, interpretations, and conclusions are the responsibility of the authors.

\begin{ack}
This work was supported by the \hyperlink{https://www.aicentre.dk/}{Pioneer Centre for Artificial Intelligence}, Denmark and the Carlsberg Foundation (\hyperlink{https://www.carlsbergfondet.dk/en/what-we-have-funded/cf25-1691/}{CF25-1691}).
\end{ack}

\printbibliography

\end{document}

%% file: figures/fail_illustration.tex
\begin{figure*}[t]
    \centering
    \includegraphics[width=0.9\linewidth]{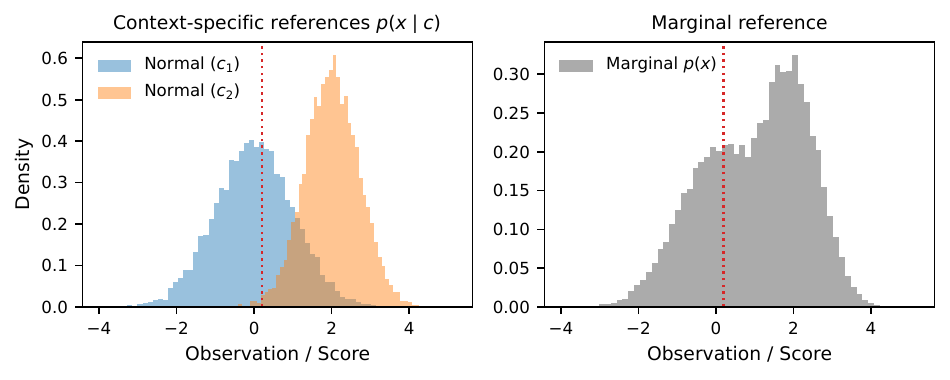}
    \caption{
    \textbf{Context-dependent versus marginal abnormality.}
    The red dotted line denotes an observation that is typical under context $c_1$ 
    but lies in the low-density tail of $p(x \mid c_2)$. 
    When contextual information is ignored and a symmetric marginal reference 
    $p(x) = \int p(x \mid c)p(c)\,dc$ is used, the same observation appears normal. 
    Marginal modeling therefore entangles contextual variability with abnormal behavior, 
    masking context-dependent anomalies.
    }
    \label{fig:context_vs_marginal}
\end{figure*}

%% file: figures/context_example.tex
\begin{figure*}[t]
    \centering
    \includegraphics[width=0.9\linewidth]{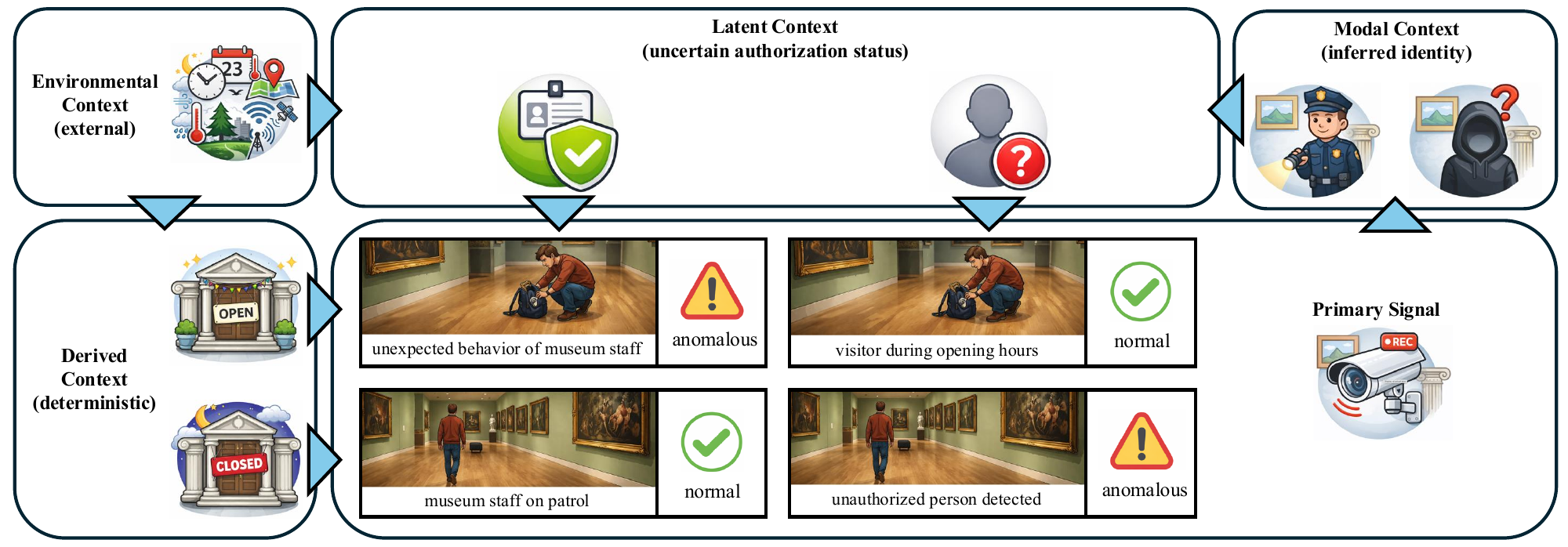}
\caption{
\textbf{Context-dependent interpretation of identical observations.}
A single visual input (person in a museum gallery) can yield different anomaly assessments depending on contextual variables with distinct observability and dependency structure. Environmental context (e.g., time) determines the operational state (open vs.\ closed) via policy (derived context). Identity constitutes modal context and is inferred from the visual signal with uncertainty. Authorization constitutes latent context and is inferred from identity, operational state, and policy. Consequently, identical actions can lead to opposite outcomes: a staff member on patrol at night is normal, whereas an unknown individual is anomalous; similarly, a visitor kneeling during opening hours is normal, but the same behavior by staff is suspicious.
}
    \label{fig:context_example}
\end{figure*}